\documentclass[journal]{IEEEtran}
\usepackage{amsmath,amsfonts}
\usepackage{algorithmic}
\usepackage{algorithm}
\usepackage{array}
\usepackage[caption=false,font=normalsize,labelfont=sf,textfont=sf]{subfig}
\usepackage{textcomp}
\usepackage{xcolor}
\usepackage{stfloats}
\usepackage{url}
\usepackage{verbatim}
\usepackage{graphicx}
\usepackage{cite}
\usepackage{nicematrix}
\usepackage{makecell}
\usepackage{booktabs}
\usepackage{multirow}
\usepackage[colorlinks=true, linkcolor=blue, citecolor=green, urlcolor=cyan]{hyperref}
\usepackage{tabularx}

\begin{document}

\title{Prompt-Consistent Cross-View Object Geo-Localization}

\author{Lingyun Xing, Wenning Hao, Kai Jiang, Dawei Jin, Xiaohan Zhang, Xiaoxiong Zhang
	\thanks{This work was supported in part by the Defense Industrial Technology Development Program under Grant JCKY2020601B018. (Corresponding author: Wenning Hao, e-mail: hwnbox@qq.com.)}
	\thanks{Lingyun Xing, Wenning Hao, Kai Jiang, and Dawei Jin are with the College of Unmanned Systems \& Intelligent Engineering, Army Engineering University of PLA, Nanjing 210007, China (e-mail: xly1231@aeu.edu.cn; hwnbox@qq.com; xdjiangkai@foxmail.com; dave\_nj@aeu.edu.cn).}
	\thanks{Xiaohan Zhang is with the College of Information Science and Electronic Engineering, Zhejiang University, Hangzhou 310058, China (e-mail: zhangxh2023@zju.edu.cn).}
	\thanks{Xiaoxiong Zhang is with the Sixty-third Research Institute, National University of Defense Technology, Nanjing 210007, China (e-mail: xiaoxiongzhang@nudt.edu.cn).}
}

\maketitle

\begin{abstract}
Cross-view object geo-localization (CVOGL) aims to locate an object indicated by a click point in a ground or drone query image within a satellite reference image. Through a quantitative prompt-sensitivity evaluation, we find that existing CVOGL methods suffer from a prompt-disobeying overfitting problem: when the click point is moved or removed, their predictions barely respond, indicating that the models rely on spurious feature correlations rather than genuinely following the prompt semantics. We trace this behavior to the limited scale of the benchmark dataset, which contains only 4,343 prompt-object pairs per task, so that the models simply memorize the object associated with each scene instead of learning the true prompt-to-object correspondence. To address this, we propose two plug-and-play components. First, we construct the CVOGL-MO dataset by expanding the existing training and test sets with diverse multi-object annotations, mitigating the memorization that induces shortcut learning. Second, we propose the Cross-View Co-Scene Multi-Object Contrastive Loss (CVMCL), which enforces distinct feature representations for different click points within the same scene, compelling the model to follow prompt instructions. We further design a prompt-sensitivity evaluation protocol with prediction-shift metrics, providing the first quantitative measurement of prompt-following capability for CVOGL. Both components are architecture-agnostic and can be integrated into existing methods without any architectural modification. Experiments on three representative backbones show that our components substantially improve prompt-following accuracy on the new multi-object test sets and markedly increase prompt sensitivity. Our dataset and code will be publicly released at https://github.com/Sunny-Fresher/CVOGL-MO.
\end{abstract}


\begin{figure}[t]
	\centering
	\includegraphics[width=\linewidth]{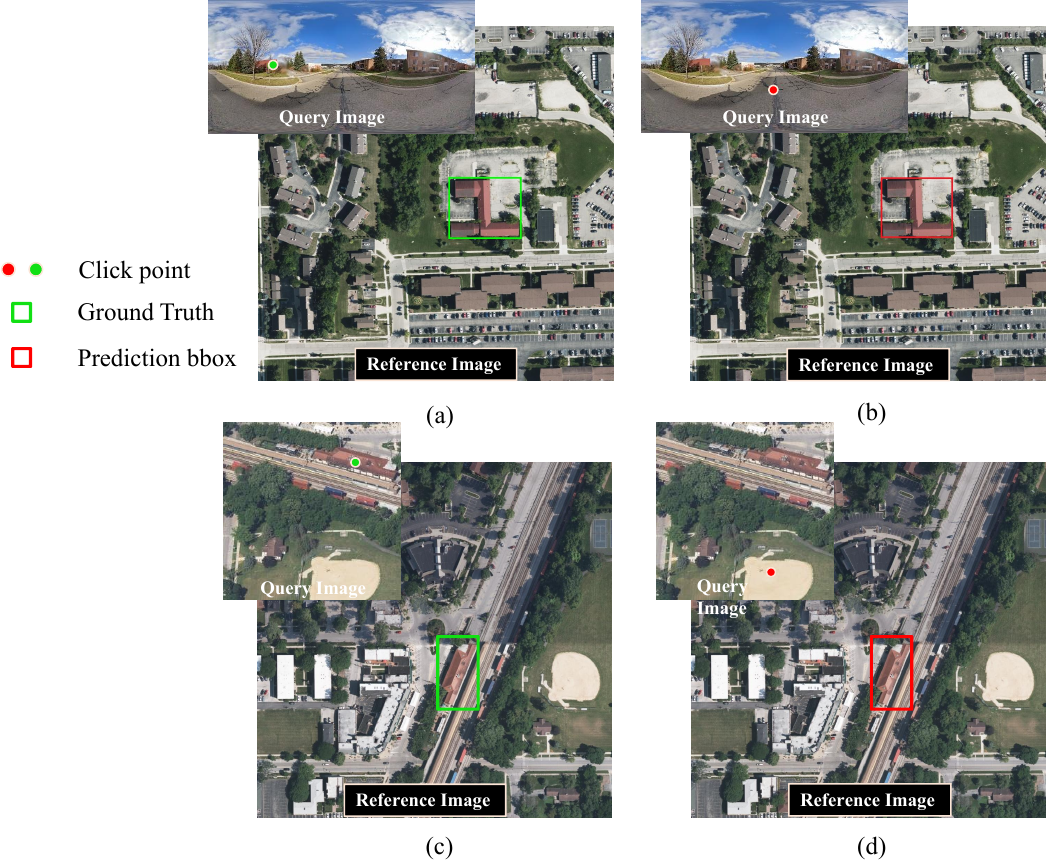}
	\caption{Illustration of the prompt-disobeying overfitting problem. (a)(b) Ground-view and (c)(d) drone-view query examples. In each case, when the click point is shifted to a different place within the same scene, existing methods persistently predict the originally annotated object instead of following the updated prompt.}
	\label{fig1}
\end{figure}

\section{Introduction}
\IEEEPARstart{C}{ROSS-VIEW} object geo-localization (CVOGL) is a pivotal task in computer vision and remote sensing, with extensive real-world applications spanning smart city management \cite{ref2, ref3, ref4}, disaster monitoring \cite{ref5, ref6}, and drone navigation \cite{ref7, ref8, ref9}. The core goal of CVOGL is to pinpoint the geographic location of a specific object indicated by a point prompt in a query image (ground-view or drone-view) within a high-resolution satellite reference image \cite{ref1}. Its development provides critical technical support for geospatial intelligence and autonomous systems \cite{ref10, ref11}, attracting sustained research attention.

In recent years, research on cross-view geo-localization has advanced rapidly, with cross-view image geo-localization (CVIGL) methods achieving notable success in addressing view gap challenges via metric learning and geometry-based strategies. These methods achieve global geographic localization for the entire reference scene by framing cross-view geo-localization as an image retrieval paradigm. However, with the growing demand for fine-grained and high-precision localization, image-level geo-localization can no longer meet such requirements. Therefore, CVOGL approaches have emerged to tackle the more demanding object-level localization task, moving beyond the image-level localization capability of CVIGL. State-of-the-art CVOGL methods, including VAGeo \cite{ref15}, OCGNet \cite{ref12}, and HALO \cite{ref14}, have effectively improved the accuracy of CVOGL and laid a foundation for subsequent research on prompt-guided object geo-localization.

However, despite the promising performance of these state-of-the-art methods, we find that they suffer from an overfitting problem that fails to truly follow point prompt instructions in essence. Specifically, these detection frameworks rely on spurious correlations in the training data rather than accurately understanding the semantic intent of the point prompt to localize the target object. This issue leads to the model making correct predictions in benchmark scenarios but failing to generalize to those scenes where prompts move to other places. This phenomenon is clearly illustrated in Fig.~\ref{fig1}. Figure (a)(b) shows the ground-view query images. When the click point is relocated from the green-marked house to the red-marked road, the model still predicts the house initially indicated by the green point and fails to adapt to the updated prompt. A similar phenomenon can be observed in the drone-view query images in Figure (c)(d): the predicted bounding box remains anchored to the object specified by the original click, even after the click position is moved. This observation suggests that the model learns spurious feature correlations instead of faithfully following the semantic instruction of the click prompt.
Beyond this qualitative observation, we quantitatively confirm this behavior. Completely removing the click prompt at test time changes the accuracy of existing models by about 1 percentage point on the new multi-object G$\rightarrow$S test set (Table~\ref{tab:prompt_cond_new}), and their predicted boxes remain almost unmoved when the click is shifted to a different object (Table~\ref{tab:prompt_shift}). This prompt-independence serves as direct evidence that existing models largely disregard the prompt and instead rely on rote memorization of the object annotated in each scene.

We identify that the root cause of this overfitting problem lies in the limited scale of the existing CVOGL benchmark dataset, which contains only 4,343 prompt-object pairs per task. With such scarce training data, the models are prone to overfit the training scenes: they simply memorize the location of the object annotated in each scene and associate it with any click point, rather than learning the true semantic correspondence between the point prompt and the indicated target. Consequently, when the click point is moved to a different object or removed entirely, the models still output the memorized object, and their predictions barely respond to the prompt change. Notably, existing CVOGL methods have focused predominantly on optimizing feature aggregation and prompt embedding mechanisms, while the critical issue of data scarcity and its resulting overfitting has not been explicitly studied.

To achieve prompt-consistent cross-view object geo-localization, we propose two complementary and plug-and-play components. First, we construct the CVOGL-MO dataset by re-annotating the training set of the benchmark CVOGL dataset with diverse multi-object annotations. The re-annotated dataset provides multiple annotated objects per scene, effectively enlarging the training data and ensuring that the model can learn the true click-to-object correspondence rather than memorizing a single fixed answer for each scene, thus avoiding ``cheating'' during training. Second, to further prevent the model from taking shortcuts \cite{ref45}, we propose the Cross-View Co-Scene Multi-Object Contrastive Loss (CVMCL), a lightweight auxiliary loss that enhances the model's ability to align prompt semantics with target objects, effectively constraining the model to focus on the true intent of the prompts. Furthermore, we construct new multi-object test sets where multiple objects are annotated per scene, enabling a more rigorous evaluation of prompt-following capability beyond the limitations of the original single-object test set. Importantly, both components are architecture-agnostic and can be directly integrated into any existing CVOGL model without modifying its backbone, fusion modules, or detection heads.

The main contributions of this work are as follows.
\begin{itemize}
\item[$\bullet$] We identify and characterize the prompt-disobeying overfitting problem in existing CVOGL methods, and propose the first quantitative prompt-sensitivity evaluation protocol for CVOGL, with prediction-shift metrics (SelfIoU, TSR and PCS) that measure whether models genuinely follow click prompts.
\item[$\bullet$] We re-annotate the training set of the benchmark dataset, constructing a more diverse prompt-target pair dataset (CVOGL-MO), which mitigates the memorization issue induced by data scarcity.
\item[$\bullet$] We propose the Cross-View Co-Scene Multi-Object Contrastive Loss (CVMCL) to constrain the model to follow prompt instructions, preventing the model from ``cheating'' by relying on spurious correlations.
\item[$\bullet$] We conduct extensive experiments on the standard CVOGL benchmark and the newly constructed multi-object test sets, verifying that our dataset and loss improve the prompt-following capability of existing methods without any architectural modification.
\end{itemize}

\section{Related Work}
\subsection{Cross-View Image Geo-Localization}
CVIGL, a pivotal task in image-level geo-localization, aims to pinpoint the geographic position of the camera that captures a ground-view query image. This goal is accomplished by establishing correspondences between the query image and the most relevant reference image \cite{ref17, ref18, ref19, ref20}, or alternatively, the most matching position within a reference image \cite{ref21, ref22, ref23}. Owing to its broad utility in fields such as autonomous navigation, aerial reconnaissance, and environmental management, CVIGL has emerged as a prominent research topic at the intersection of computer vision and geographic information science.

Existing research on CVIGL can be divided into two primary categories: metric learning-based methods and geometry-based methods \cite{ref1}. For metric learning-based methods \cite{ref24, ref25, ref26, ref27, ref28, ref29, ref30, ref31, ref32, ref33}, the primary focus lies in learning features that are invariant to changes in viewpoint, with little consideration given to geometric structures. Recent advancements in this area have involved the integration of self-distillation techniques, multi-scale feature reasoning \cite{ref34}, and multi-scale attention mechanisms \cite{ref35}. These enhancements facilitate improved cross-view matching by enabling hierarchical feature interaction and self-supervised guidance. Furthermore, metric learning-based methods have been adapted to UAV-to-satellite geo-localization scenarios \cite{ref36}, shedding light on the domain shift induced by differences in altitude and scale. Additionally, diffusion-based architectures have been employed to tackle extreme environmental variations \cite{ref37}, thereby broadening the scope of application for this class of methods.

On the other hand, geometry-based methods \cite{ref20, ref38, ref39, ref40, ref41, ref42} aim to narrow the viewpoint gap between ground-view and aerial-view images explicitly through the utilization of geometric constraints, such as orientation data, which in turn enables precise camera geo-localization. These methods underscore the critical role of geometric information in addressing the cross-view discrepancy, a major bottleneck in CVIGL research.

While substantial advancements have been made in camera-level localization using existing CVIGL techniques, notable limitations remain. In particular, current CVIGL methods are confined to camera-level localization and are unable to achieve precise object-level localization \cite{ref1}. In other words, these approaches can only estimate the approximate location of the query image relative to the reference image, which fails to satisfy the demands of object-level geo-localization tasks.

\subsection{Cross-View Object Geo-Localization}
Cross-View Object Geo-Localization (CVOGL) is an object-level task that locates an object indicated by a point prompt in a query image (ground-view or drone-view) on reference satellite imagery. DetGeo \cite{ref1} first formalized this task and proposed a detection-based framework to address the limitation of image-level CVIGL methods.

Subsequent works enhanced performance by tackling core challenges. OCGNet \cite{ref12} adopted Gaussian Kernel Transfer (GKT) for precise prompt embedding, a late-stage Location Enhancement (LE) module, and Multi-Head Cross Attention (MHCA) to handle similar objects; GeoFormer \cite{ref13} incorporates SAM's knowledge into the network by using it as part of the prompt encoder and applying knowledge distillation, and is the first to generate accurate object-level prompts for CVOGL with a large-scale model; HALO \cite{ref14} proposed a cascade cross-view interaction module (CCIM) and adaptive feature hierarchization (AFHAM), combined with contrastive learning and hard sample mining; VAGeo \cite{ref15} introduced view-specific positional encoding (VSPE) and channel-spatial hybrid attention (CSHA) to adapt to different query viewpoints and strengthen discriminative features; ReCOT \cite{ref16} reformulated CVOGL as a recurrent task with learnable tokens, supplemented by SAM-based knowledge distillation and hierarchical attention for reference feature enhancement, achieving state-of-the-art results with fewer parameters.

These advances continuously improve the robustness of CVOGL to viewpoint discrepancies and prompt ambiguity, and several of them explicitly enhance prompt encoding and object distinguishing \cite{ref13, ref16}. However, these efforts all optimize the model side, while the limited scale of the benchmark dataset---which contains only 4,343 prompt-object pairs with a single annotated object per scene---and the resulting prompt-disobeying overfitting have not been explicitly studied. Our work is complementary to model-side improvements. We diagnose this data-side problem quantitatively and address it with data-level and optimization-level components that can be plugged into any of the above architectures.

\section{Dataset Construction}
To mitigate data scarcity and the resulting prompt-disobeying overfitting in cross-view object geo-localization, we construct the CVOGL-MO dataset, which expands the original CVOGL dataset with additional multi-object annotations on existing query-reference pairs. This data optimization enlarges the training set and provides fundamental spatial correspondence prior for our subsequent prompt-following constraint loss.

\subsection{Limitations of the Original CVOGL Dataset}
As revealed in previous analysis, the existing cross-view object geo-localization dataset is limited in scale, containing only 4,343 prompt-object pairs per task, with each scene paired with a single annotated object. With such scarce training data, the models are prone to overfit and simply memorize the object annotated in each scene rather than learning the true click-to-object correspondence. This induces severe prompt-disobeying overfitting for current models. When the input click prompt is altered, the model still tends to locate the previously memorized object instead of following the updated click instruction, resulting in undesirable shortcut learning and degraded prompt-following ability.

\begin{figure}[t]
\centering
\includegraphics[width=\linewidth]{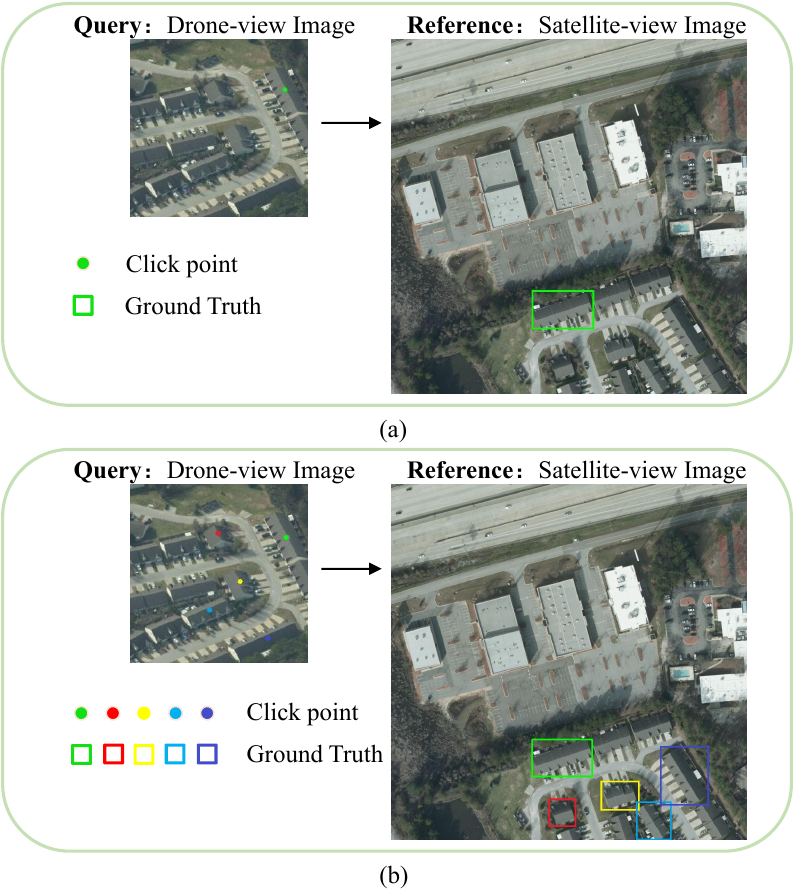}
\caption{Comparison of the CVOGL dataset and our CVOGL-MO dataset.
(a) The CVOGL dataset: A single query click point (green dot) corresponds to one matched bounding box (green box) in the satellite-view reference image.
(b) Our extended dataset: We augment the original CVOGL dataset with additional query click points (yellow, blue, red, and purple dots) and their corresponding multi-target bounding boxes (yellow, blue, red, and purple boxes) in the satellite-view image, while retaining the original green point-box pair from the CVOGL dataset for consistency.}
\label{fig2}
\end{figure}

\subsection{Annotation Principles and Expansion Strategy}
To alleviate the aforementioned data scarcity and suppress model overfitting, we conduct systematic dataset expansion based on the original benchmark. Based on existing click-point annotations and the corresponding bounding boxes, we further annotate additional corresponding object pairs on identical query-reference image groups, and integrate these newly annotated samples into the official annotation file.

\textbf{Object selection criteria.} Beyond the originally annotated object, we supplement annotations for objects that are clearly identifiable in both the query and the satellite views, giving priority to three under-represented categories: (1) inconspicuous objects with weak visual signatures; (2) small instances occupying less than 1\% of the reference image area; and (3) targets with mild occlusion. Objects that are severely occluded, ambiguous in extent, or not verifiable in the satellite view are excluded. Each image pair is allowed to contain at most 8 annotated objects.

\textbf{Boundary conventions.} To ensure a consistent prompt-to-box mapping, we adopt explicit boundary rules for ambiguous object types. For sports facilities, each individual sub-court (e.g., a single basketball court) is annotated as a separate object rather than boxing the entire court complex, and the click point is placed at the center of the corresponding sub-court. For irregularly shaped building complexes (e.g., gate-shaped or L-shaped structures), the bounding box encloses the complete connected structure.

\textbf{Consistency verification.} All newly added annotations are independently checked by a senior reviewer, and disagreements are resolved by discussion.

Through this expansion strategy, we effectively enlarge the training data, preventing the model from simply memorizing a single object per scene and encouraging it to learn genuine click-target spatial correspondence. The comparison of the CVOGL dataset and our proposed dataset can be seen in Fig.~\ref{fig2}.

In addition to expanding the training set, we further construct new multi-object test sets to more rigorously evaluate the prompt-following capability of CVOGL models. Specifically, we select subsets of the original test set and annotate multiple objects per image. We select scenes in which at least two additional objects satisfying the above criteria can be verified in both views and the selection is made before any model evaluation and is not influenced by model predictions.

These new test sets enable a more comprehensive evaluation of whether models can truly follow different click prompts on the same scene to locate different objects. We note that the new test sets share query--reference image pairs with the original benchmark. They are designed to measure prompt-following capability within familiar scenes rather than generalization to unseen scenes.

 \begin{table}[htbp]
 	\centering
 	\caption{Dataset Statistics for CVOGL and CVOGL-MO}
 	\label{tab:dataset}
 	\renewcommand{\arraystretch}{1.5}
 	\scriptsize
 	\begin{tabular}{c|c|c|c|c}
 		\hline
 		Task & Split & Instances & Query & Reference \\
 		\hline\hline
 		\multirow{5}{*}{Ground $\rightarrow$ Satellite}
 		& training (CVOGL)     & 4,343 & 3,674 & 4,063 \\
 		& training (CVOGL-MO) & 8,749 & 3,674 & 4,063 \\ 		& validation (CVOGL)  & 923   & 923   & 923   \\
 		& test (CVOGL)      & 973   & 812   & 910   \\
 		& new test (multi-obj) & 828   & 343   & 343   \\ 
 		\hline
 		\multirow{5}{*}{Drone $\rightarrow$ Satellite}
 		& training (CVOGL)     & 4,343 & 3,674 & 4,063 \\
 		& training (CVOGL-MO) & 9,452 & 3,674 & 4,063 \\ 
 		& validation (CVOGL)  & 923   & 923   & 923   \\
 		& test (CVOGL)      & 973   & 812   & 910   \\
 		& new test (multi-obj) & 1,091 & 450   & 450   \\ 
 		\hline
 	\end{tabular}
 \end{table}

\subsection{Statistical Properties}
Based on the original CVOGL dataset consisting of 4,343 prompt-object pairs for both the ground to satellite (G$\rightarrow$S) and drone to satellite (D$\rightarrow$S) tasks, we perform extensive supplementary annotation. After expansion, the total number of annotated training instances reaches 9,452 for the D$\rightarrow$S task and 8,749 for the G$\rightarrow$S task. The G$\rightarrow$S new test set contains 343 images with 828 annotated instances, while the D$\rightarrow$S new test set contains 450 images with 1,091 annotated instances. Detailed statistical comparisons between the original and augmented datasets are summarized in Table~\ref{tab:dataset}.

The two tasks cover the same 3,674 training scenes as the original benchmark. The instance counts differ because the number of additionally annotated objects per scene differs between tasks. Ground panoramic images are usually dominated by ground regions, leaving relatively few objects available for annotation. In contrast, the UAV viewpoint bears greater similarity to the satellite viewpoint and contains more targets to be annotated.

\begin{figure}[t]
	\centering
	\includegraphics[width=\columnwidth]{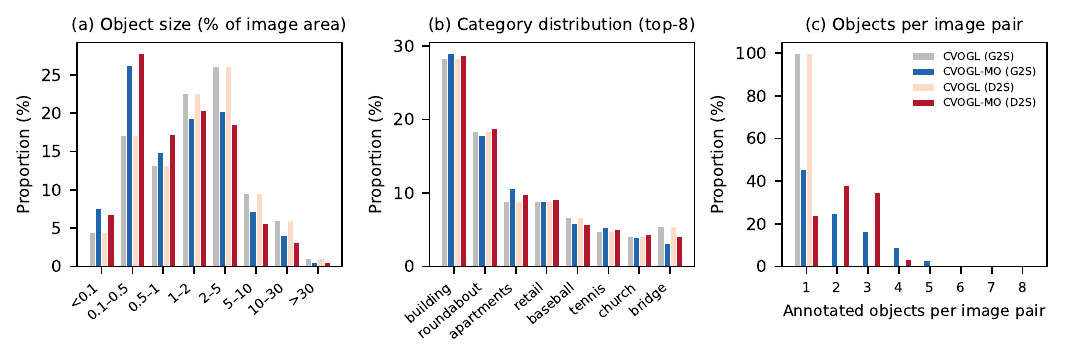}
	\caption{Statistical comparison of the original CVOGL dataset and our CVOGL-MO dataset on the training split: distributions of (a) object size (percentage of image area), (b) object category (top-8), and (c) the number of annotated objects per image pair.}
	\label{fig:dataset_stats}
\end{figure}

Fig.~\ref{fig:dataset_stats} further compares the distributions of object size, category, and the number of annotated objects per image pair before and after expansion. After expansion, the proportion of small objects (occupying less than 1\% of the image area) increases from 34.7\% to 48.7\% for the G$\rightarrow$S task and to 51.9\% for the D$\rightarrow$S task, and the category distribution becomes more balanced.

\begin{figure*}[htbp]
  \centering
  \includegraphics[width=\textwidth]{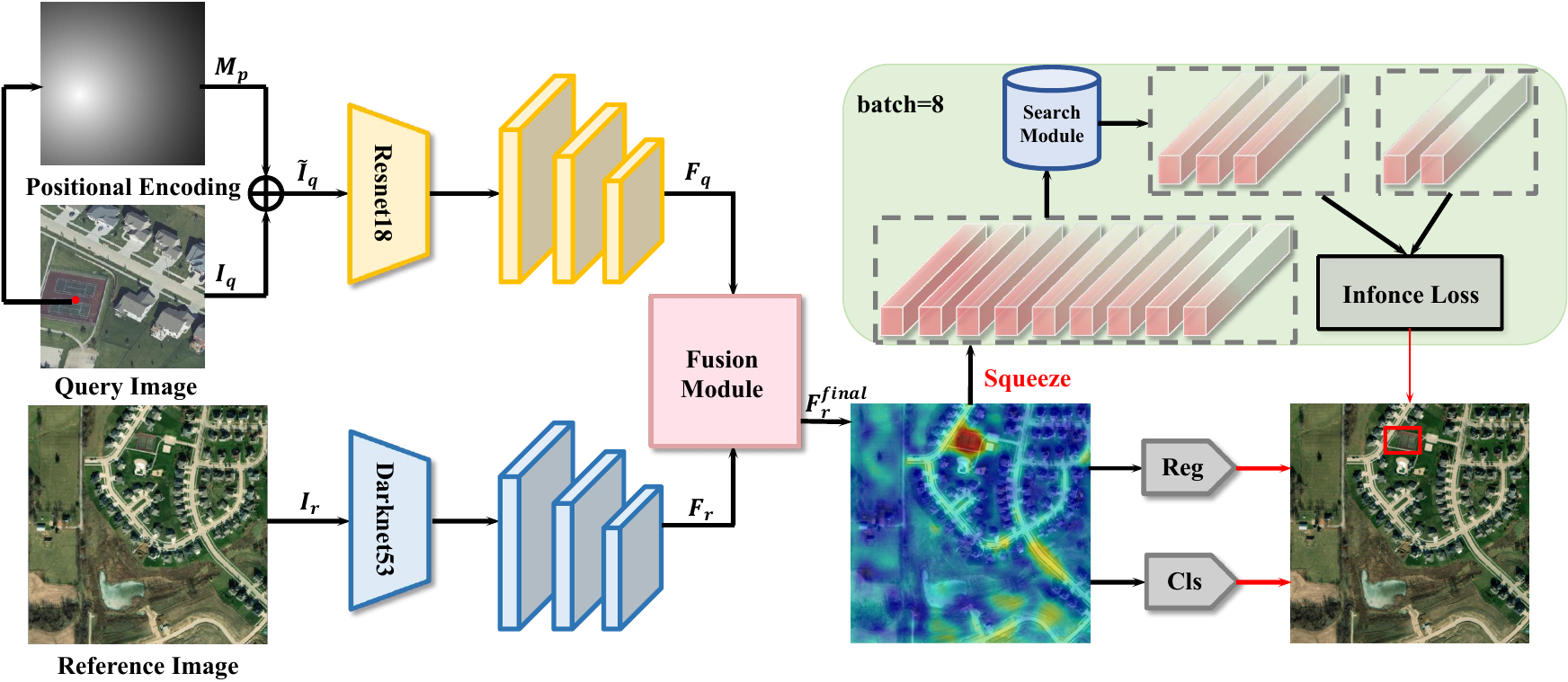}
  \caption{Overview of our approach. The encoder and feature fusion components follow the standard CVOGL paradigm (DetGeo \cite{ref1}). Our proposed CVMCL operates on the fused features as a plug-and-play auxiliary loss, requiring no modification to existing architectures.}
  \label{fig3}
\end{figure*}

\section{Methodology}
\label{sec:method}

In this section, we first introduce the overall pipeline of the backbone used in baseline approaches, in order to provide readers with a clear understanding of the common workflow for the CVOGL task. We then elaborate on our proposed CVMCL in detail. Our method does not modify the backbone architectures or basic detection pipelines of existing models. All original network structures, including feature encoders, fusion modules, and detection heads, are fully preserved.

\subsection{Overall Pipeline of Baseline Backbones}
Given a query image $I_q$ (ground-view or drone-view), a user click point $p=(x,y)$ that indicates the target object, and a satellite reference image $I_r$, the goal is to estimate the bounding box $\mathcal{B}$ of the corresponding object in $I_r$.

The pipeline follows a standard encode-fuse-detect workflow:
\begin{itemize}
    \item The click point is embedded into a spatial prompt map and fused with the query image.
    \item The enhanced query image is encoded by a query encoder to extract multi-scale features.
    \item The satellite reference image is encoded by a reference encoder to obtain spatial and semantic features.
    \item Cross-view feature fusion aligns query and reference features via similarity learning and attention.
    \item The fused reference features are fed into the original detection head to locate the target.
\end{itemize}

\subsubsection{Click Prompt Embedding}
To inject user intent into the model, we encode the click point into a spatial attention map $M_p$. This map highlights the target region in the query image and is concatenated with the original image as an additional channel:
\begin{equation}
    \widetilde{I}_q = I_q \oplus M_p
\end{equation}
where $\oplus$ denotes channel-wise concatenation.

\subsubsection{Query Image Encoder}
The enhanced query image $\widetilde{I}_q$ is fed into a query encoder $\mathcal{E}_q(\cdot)$ to extract multi-level visual features:
\begin{equation}
    F_q = \mathcal{E}_q(\widetilde{I}_q)
\end{equation}
The encoder outputs a pyramid of features that preserve both low-level spatial details and high-level semantics.

\subsubsection{Reference Image Encoder}
The satellite reference image $I_r$ is processed by a reference encoder $\mathcal{E}_r(\cdot)$ to produce multi-scale spatial features:
\begin{equation}
    F_r = \mathcal{E}_r(I_r)
\end{equation}
This branch focuses on capturing layout, structure, and spatial distribution in the satellite image.

\subsubsection{Cross-View Feature Fusion}
To align features from ground/drone and satellite views, the pipeline adopts a cross-view fusion module. First, a normalized similarity matrix between $F_q$ and $F_r$ is computed:
\begin{equation}
    \mathcal{S} = \text{Norm}(F_q^\top F_r)
\end{equation}
where $\text{Norm}(\cdot)$ denotes normalization.

Then, a cross-view attention mechanism is applied to weight the reference features:
\begin{equation}
    F_r^\text{aligned} = \mathcal{S} \cdot F_r
\end{equation}
where $\mathcal{S}$ represents attention weights derived from cross-view similarity.

Finally, the aligned reference features are fused via a convolution block:
\begin{equation}
    F_r^\text{final} = \text{CBR}(F_r^\text{aligned})
\end{equation}
where $\text{CBR}(\cdot)$ denotes the Conv-BN-ReLU operation.

These fused features are directly sent into the original detection head of the baseline model to generate the final bounding box prediction.

\subsection{Cross-View Co-Scene Multi-Object Contrastive Loss}
\subsubsection{Design Motivation}
Conventional cross-view object geo-localization models are optimized only using regression loss for bounding box prediction and confidence loss for object existence classification. Such supervision guides the model to complete basic localization tasks, but cannot ensure that the network learns the genuine spatial correspondence between click points and target objects. In practice, models trained with these losses on scarce data tend to memorize the object annotated in each scene, regardless of the position of the input click point. Even when the click point is manually moved to another object, the model still predicts the originally annotated object, indicating that it memorizes the fixed answer rather than learning the correct point-object mapping.

To address this critical issue, we propose the Cross-View Co-Scene Multi-Object Contrastive Loss (CVMCL). By imposing contrastive constraints on multiple objects within the same scene, our loss explicitly enlarges the feature discrepancy between different objects activated by different click points on the same query-reference pair. In this way, the model is forced to learn accurate one-to-one correspondence between click positions and their corresponding object instances, instead of simply memorizing fixed outputs.

\subsubsection{Co-Scene Instance Grouping in Mini-Batches}
During training, each mini-batch is denoted as $\mathcal{M} = \{(\boldsymbol{q}_i, \boldsymbol{r}_i, \boldsymbol{p}_i, \boldsymbol{b}_i, ID_i)\}_{i=1}^B$, where $B$ is the batch size and $ID_i$ represents the unique scene identifier corresponding to each cross-view image pair. Samples that share the same $ID_i$ originate from an identical query-reference pair and thus belong to the same co-scene multi-object group. We formally organize such samples into the $k$-th group $\mathcal{G}_k$, defined as:

\begin{equation}
\label{equ:group}
\mathcal{G}_k = \{(\boldsymbol{q}_{1}, \boldsymbol{r}_{1}, \boldsymbol{p}_{1}, \boldsymbol{b}_{1}), \dots, (\boldsymbol{q}_{n_k}, \boldsymbol{r}_{n_k}, \boldsymbol{p}_{n_k}, \boldsymbol{b}_{n_k})\}
\end{equation}

where $n_k$ denotes the number of samples contained in the $k$-th group. Notably, even if no repeated scene $ID$ exists in the current batch, all instances are directly regarded as a single global group for contrastive learning. When the batch contains multiple co-scene groups (i.e., several subsets of samples each sharing a distinct scene $ID$), the contrastive loss is first computed within each group via Eq.~(\ref{equ4})--(\ref{equ5}), and the resulting per-group losses are averaged to obtain the final CVMCL in Eq.~(\ref{equ6}). By confining contrastive learning within each scene group, the model is forced to discriminate objects based on prompt semantics rather than trivial scene-level differences.

\textbf{Mini-batch construction.} Since co-scene groups rarely co-occur under uniform random sampling with a small batch size, we employ a scene-aware batch sampler. Concretely, samples are first grouped by their scene identifier $ID_i$; the resulting groups are shuffled and flattened into a single sampling sequence so that samples from the same scene appear consecutively, and the sequence is then partitioned into fixed-size mini-batches of size $B$, with the last incomplete batch discarded to avoid undersized chunks that cause BatchNorm instability under data parallelism. With this sampler, each mini-batch contains on average about 2.5 co-scene groups (for the default batch size of 8), ensuring that the contrastive objective is effectively triggered in every training step.

\subsubsection{Feature Embedding and Loss Computation}
For each sample, the high-level fused feature map $\boldsymbol{X}_i \in \mathbb{R}^{C \times H \times W}$ output by the cross-view fusion module encodes spatial and channel-wise structural information. We compress it into a compact feature vector via global average pooling (GAP):

\begin{equation}
	\label{equ1}
	\boldsymbol{z}_i = \text{GAP}(\boldsymbol{X}_i) = \frac{1}{H W} \sum_{h=1}^{H} \sum_{w=1}^{W} \boldsymbol{X}_{i}(:,h,w)
\end{equation}

where $\boldsymbol{z}_i \in \mathbb{R}^C$ denotes the $C$-dimensional global feature vector of the $i$-th sample in the group. To remove the effect of feature magnitude, we L2-normalize each embedding:
\begin{equation}
	\label{equ2}
	\tilde{\boldsymbol{z}}_i = \frac{\boldsymbol{z}_i}{\|\boldsymbol{z}_i\|_2}
\end{equation}

The cosine similarity between normalized features of sample $i$ and sample $j$ is calculated as:
\begin{equation}
	\label{equ3}
	s_{i,j} = \text{sim}(\tilde{\boldsymbol{z}}_i, \tilde{\boldsymbol{z}}_j) = \tilde{\boldsymbol{z}}_i^\top \tilde{\boldsymbol{z}}_j
\end{equation}

Following the standard InfoNCE paradigm, the contrastive loss for the $i$-th query sample is defined as:
\begin{equation}
	\label{equ4}
	\ell_i = -\log \frac{\exp\left(s_{i,i} / \tau\right)}{\sum_{j=1}^{n_k} \exp\left(s_{i,j} / \tau\right)}
\end{equation}

The group-level contrastive loss is then obtained by averaging the per-sample losses over all $n_k$ instances within the group:
\begin{equation}
	\label{equ5}
	\mathcal{L}_{\text{cvmc}} = \frac{1}{n_k}\sum_{i=1}^{n_k} \ell_i
\end{equation}

where $n_k$ denotes the number of instances in the group. Within a co-scene group, each instance corresponds to a different click prompt, so no two instances form a natural positive pair. The loss separates them, preventing the model from mapping distinct prompts to the same output. Since the embeddings are L2-normalized, the self-similarity term $s_{i,i}=1$ is a constant that serves only as an anchor, so the loss reduces to pushing all distinct instances within the group apart, consistent with the absence of any positive pair.

Finally, the total CVMCL for the entire batch is obtained by averaging the group-level losses over all $K$ groups:

\begin{equation}
\label{equ6}
\mathcal{L}_{\text{cvmc}}^{\text{batch}} = \frac{1}{K} \sum_{k=1}^{K} \mathcal{L}_{\text{cvmc}}^{(k)}
\end{equation}

\subsection{Total Training Objective}
The final training objective integrates three complementary loss terms in a weighted sum:

\begin{equation}
\label{equ7}
\mathcal{L}_{\text{total}} = \mathcal{L}_{\text{cls}} + \beta \mathcal{L}_{\text{geo}} + \gamma \mathcal{L}_{\text{cvmc}}^{\text{batch}}
\end{equation}

where $\mathcal{L}_{\text{cls}}$ denotes the classification loss for predicting the confidence score of the target object, and $\mathcal{L}_{\text{geo}}$ denotes the localization loss for bounding box regression. Following the original DetGeo\cite{ref1} setup, we fix $\beta=1$ to maintain consistency with the basic optimization objective. $\gamma$ is a balancing weight that controls the contribution of our proposed CVMCL. The three loss terms jointly optimize the model that classification and localization losses guarantee the basic object detection capability, while the contrastive loss strengthens intra-scene feature discrimination and suppresses the memorization of a single object per scene, encouraging the model to truly learn accurate prompt consistency.

\section{Experiments}
\label{sec:experiments}

\subsection{Experimental Settings}
\subsubsection{Implementation Details}
The model is implemented in PyTorch and the training process is conducted on 2 NVIDIA V100 GPUs. We use the RMSProp optimizer with an initial learning rate of $10^{-4}$, which decays by half every 10 epochs. We conduct experiments on three existing models which have open source code to verify the effectiveness of our method. For DetGeo, the batch size is set to 8 and the model is trained for 30 epochs. For HALO, we adopt a batch size of 16 and train for 40 epochs. For OCGNet, we use a batch size of 8 and train for 30 epochs for the ground to satellite task while a batch size of 12 and train for 40 epochs for the drone to satellite task. The balance weight $\gamma$ in CVMCL is set to $1\times 10^{-4}$ and the temperature coefficient $\tau$ is set to $0.1$. All data augmentation strategies, backbone initialization, and pre-training settings are strictly aligned with the baseline methods to ensure experimental validity.

\subsubsection{Evaluation Metrics}
Following the pioneering work DetGeo \cite{ref1}, we adopt Intersection over Union (IoU) and IoU-based accuracy as the core evaluation metrics to comprehensively measure the localization precision of the model. Given a predicted bounding box $b_i$ and its corresponding ground-truth bounding box $b_i^*$, the IoU is defined as:
\begin{equation}
\text{IoU}(b_i, b_i^*) = \frac{\left| b_i \cap b_i^* \right|}{\left| b_i \cup b_i^* \right|}
\end{equation}
where $\left| b_i \cap b_i^* \right|$ denotes the intersection area between the predicted and ground-truth boxes, and $\left| b_i \cup b_i^* \right|$ denotes their union area.

The accuracy is defined as:
\begin{equation}
	\mathrm{Acc}@t = \frac{1}{n} \sum_{i=1}^{n} \psi_i(t)
\end{equation}
where
\[
\psi_i(t) =
\begin{cases}
	1, & \text{if } \mathrm{IoU}(b_i, b_i^*) \ge t \\
	0, & \text{else}
\end{cases}
\]
We adopt the settings $t = 0.25$ and $t = 0.5$ in this paper.

\subsubsection{Parameter Analysis}
\label{sec:param_analysis}
\begin{figure}[t]
	\centering
	\includegraphics[width=\columnwidth]{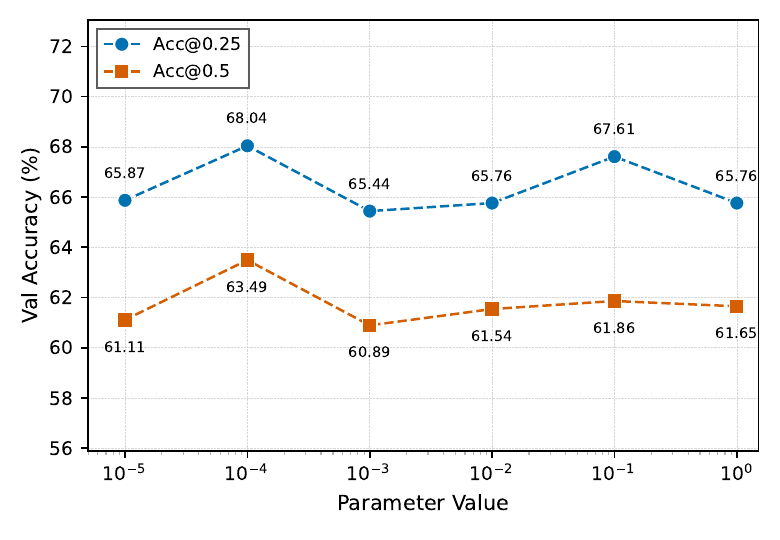}
	\caption{Validation accuracy (\%) of D$\rightarrow$S task under different values of the balance weight $\gamma$. Acc@0.25 and Acc@0.5 denote the IoU-thresholded accuracy at thresholds 0.25 and 0.5, respectively.}
	\label{fig:param_accuracy}
\end{figure}

To analyze the influence of the CVMCL balance weight $\gamma$, we train models with $\gamma$ ranging from $1$ to $1\times 10^{-5}$ and evaluate them on the validation set. As shown in Fig.~\ref{fig:param_accuracy}, both $\mathrm{Acc}@0.25$ and $\mathrm{Acc}@0.5$ of the drone-to-satellite task remain relatively stable over this wide range, with total variation not exceeding 3\% for either metric. This suggests that the model is robust to the choice of $\gamma$ and does not require extensive tuning in practice. The best validation performance is attained at $\gamma = 1\times 10^{-4}$, which we adopt for all remaining experiments; the test set is used only for the final evaluation.

\subsection{Prompt Sensitivity Evaluation}
\label{sec:prompt_sensitivity}

The central claim of this paper is that existing CVOGL models do not genuinely follow the click prompt. To substantiate this claim quantitatively, we design a prompt-sensitivity evaluation protocol that measures how model predictions respond to controlled prompt perturbations.

\subsubsection{Protocol}
We evaluate each model under four prompt conditions:
\begin{itemize}
	\item \textbf{C0 (original click):} the annotated click point, i.e., the standard setting;
	\item \textbf{C1 (shifted click):} the click is moved to another annotated object within the same scene. This condition is evaluated on the new multi-object test sets by enumerating all pairs of annotated objects per scene;
	\item \textbf{C2 (no click):} the click prompt is removed by zeroing the prompt map $M_p$.
\end{itemize}

\subsubsection{Metrics}
Under each condition $c$, we report the condition-wise accuracy Acc@$t(c)$; for C1, the ground truth of the shifted click is the bounding box of the newly indicated object. In addition, for every object pair $(A_i, A_j)$ within the same scene, let $\hat{b}^{(A_i)}$ and $\hat{b}^{(A_j)}$ denote the predicted boxes when the click is placed on $A_i$ and $A_j$, respectively. We define three prediction-shift metrics:

\textbf{Self-IoU} measures how much the prediction changes when the click moves (lower is better for prompt-following):
\begin{equation}
	\mathrm{SelfIoU} = \frac{1}{M} \sum_{(i,j)} \mathrm{IoU}\left(\hat{b}^{(A_i)}, \hat{b}^{(A_j)}\right)
\end{equation}

\textbf{Target-Switch Rate (TSR)} measures whether the prediction after the click shift correctly lands on the newly indicated object:
\begin{equation}
	\mathrm{TSR} = \frac{1}{M} \sum_{(i,j)} \left[\mathrm{IoU}\left(\hat{b}^{(A_j)}, b_{A_j}^{*}\right) \ge 0.5\right]
\end{equation}

Finally, we combine target-following accuracy and prediction responsiveness into a single \textbf{Prompt Consistency Score (PCS)}:
\begin{equation}
	\mathrm{PCS} = \mathrm{TSR} \times \left(1 - \overline{\mathrm{SelfIoU}}\right)
\end{equation}
where $\overline{\mathrm{SelfIoU}}$ is the mean SelfIoU expressed as a fraction in $[0,1]$ (the percentage values in Table~\ref{tab:prompt_shift} are divided by 100). A model that ignores the prompt scores close to 0, while an ideal prompt-following model approaches 1.

\subsubsection{Results}
As shown in Table~\ref{tab:prompt_cond_old}, Table~\ref{tab:prompt_cond_new}, and Table~\ref{tab:prompt_shift}, the three baselines exhibit high SelfIoU on the new multi-object test sets (e.g., 75.80\%, 66.00\%, and 51.99\% on G$\rightarrow$S for DetGeo, OCGNet, and HALO, respectively), indicating that their predictions barely move when the click is shifted. Removing the click prompt (C2) changes the baseline accuracy by only about one percentage point on the new multi-object G$\rightarrow$S test set, confirming that the models largely disregard the prompt. In contrast, integrating CVOGL-MO and CVMCL consistently reduces SelfIoU and raises TSR and PCS on both tasks, demonstrating that our method compels the models to genuinely follow the click prompt. Notably, this prompt-disobeying overfitting is more pronounced for the G$\rightarrow$S task than for the D$\rightarrow$S task, because the drone and satellite views are visually more similar, the cross-view gap is smaller and the shortcut is less severe for D$\rightarrow$S. Nevertheless, after integrating our method, the prompt becomes markedly more effective on both tasks, as reflected by the reduced SelfIoU and the increased TSR and PCS.

\begin{table}[t]
	\centering
	\caption{Acc@0.5 (\%) under prompt perturbations on the old test sets. C0: original click; C2: click removed. Results are averaged over three seeds.}
	\label{tab:prompt_cond_old}
	\setlength{\tabcolsep}{3pt}
	\renewcommand{\arraystretch}{1.2}
	\begin{tabular}{l | cc | cc}
		\toprule
		\multirow{2}{*}[-0.5em]{Method}
		& \multicolumn{2}{c|}{Ground $\rightarrow$ Satellite}
		& \multicolumn{2}{c}{Drone $\rightarrow$ Satellite} \\
		\cmidrule(lr){2-3} \cmidrule(lr){4-5}
		& C0 & C2 & C0 & C2 \\
		\midrule
		DetGeo \cite{ref1}      & 42.24 & 44.81 & 57.66 & 52.31 \\
		DetGeo + Ours           & 36.38 & 38.34 & 64.65 & 38.34 \\
		\midrule
		OCGNet \cite{ref12}     & 47.69 & 46.56 & 63.93 & 52.11 \\
		OCGNet + Ours           & 43.78 & 38.44 & 64.95 & 40.08 \\
		\midrule
		HALO \cite{ref14}       & 46.76 & 46.66 & 64.44 & 54.16 \\
		HALO + Ours             & 43.06 & 40.90 & 70.20 & 44.19 \\
		\bottomrule
	\end{tabular}
\end{table}

\begin{table}[t]
	\centering
	\caption{Acc@0.5 (\%) under prompt perturbations on the new multi-object test sets. C0: original click; C2: click removed. Results are averaged over three seeds.}
	\label{tab:prompt_cond_new}
	\setlength{\tabcolsep}{3pt}
	\renewcommand{\arraystretch}{1.2}
	\begin{tabular}{l | cc | cc}
		\toprule
		\multirow{2}{*}[-0.5em]{Method}
		& \multicolumn{2}{c|}{Ground $\rightarrow$ Satellite}
		& \multicolumn{2}{c}{Drone $\rightarrow$ Satellite} \\
		\cmidrule(lr){2-3} \cmidrule(lr){4-5}
		& C0 & C2 & C0 & C2 \\
		\midrule
		DetGeo \cite{ref1}      & 20.77 & 21.74 & 25.57 & 23.01 \\
		DetGeo + Ours           & 23.91 & 20.77 & 37.40 & 18.88 \\
		\midrule
		OCGNet \cite{ref12}     & 22.71 & 21.50 & 32.08 & 22.36 \\
		OCGNet + Ours           & 27.05 & 19.20 & 41.15 & 18.79 \\
		\midrule
		HALO \cite{ref14}       & 23.79 & 22.83  & 34.10 & 24.20 \\
		HALO + Ours             & 27.17 & 20.53 & 44.18 & 22.27 \\
		\bottomrule
	\end{tabular}
\end{table}

\begin{table*}[t]
	\centering
	\caption{Prediction-shift metrics under click switching (C1) on the new multi-object test sets. $\downarrow$/$\uparrow$ indicate the desired direction.}
	\label{tab:prompt_shift}
	\setlength{\tabcolsep}{4pt}
	\renewcommand{\arraystretch}{1.2}
	\begin{tabular}{l | ccc | ccc}
		\toprule
		\multirow{2}{*}[-0.5em]{Method}
		& \multicolumn{3}{c|}{Ground $\rightarrow$ Satellite}
		& \multicolumn{3}{c}{Drone $\rightarrow$ Satellite} \\
		\cmidrule(lr){2-4} \cmidrule(lr){5-7}
		& SelfIoU$\downarrow$(\%) & TSR$\uparrow$(\%) & PCS$\uparrow$(\%)
		& SelfIoU$\downarrow$(\%) & TSR$\uparrow$(\%) & PCS$\uparrow$(\%) \\
		\midrule
		DetGeo \cite{ref1}      & 75.80 & 17.80 & 4.40 & 44.10 & 25.20 & 14.10 \\
		DetGeo + Ours           & 60.30 & 18.30 & 7.10 & 19.90 & 32.30 & 25.80 \\
		\midrule
		OCGNet \cite{ref12}     & 66.00 & 21.00 & 7.10 & 22.59 & 27.58 & 21.35 \\
		OCGNet + Ours           & 34.00 & 24.70 & 16.30 & 10.33 & 36.28 & 32.52 \\
		\midrule
		HALO \cite{ref14}       & 51.99 & 20.78 & 9.98 & 16.35 & 35.58  & 29.76 \\
		HALO + Ours             & 24.31 & 27.68 & 20.95 &10.10 & 37.40 & 33.62 \\
		\bottomrule
	\end{tabular}
\end{table*}

\begin{table}[t]
	\centering
	\caption{Performance Comparison on the New Multi-Object Test Sets (\%).}
	\label{tab:new_test}
	\setlength{\tabcolsep}{2pt}
	\renewcommand{\arraystretch}{1.2}
	\begin{tabular}{c | cc | cc}
		\toprule
		\multirow{2}{*}[-0.9em]{{Method}}
		& \multicolumn{2}{c|}{Ground $\rightarrow$ Satellite}
		& \multicolumn{2}{c}{Drone $\rightarrow$ Satellite} \\
		\cmidrule(lr){2-3} \cmidrule(lr){4-5}
		& Acc@0.25 & Acc@0.5 & Acc@0.25 & Acc@0.5 \\
		\midrule
		DetGeo \cite{ref1}      & 21.74 & 20.77 & 27.59 & 25.57 \\
		Ours (DetGeo)           & \textbf{25.24} & \textbf{23.91} & \textbf{39.60} & \textbf{37.40} \\
		\midrule
		OCGNet \cite{ref12}     & 24.03 & 22.71 & 35.01 & 32.08 \\
		Ours (OCGNet)           & \textbf{29.35} & \textbf{27.05} & \textbf{44.00} & \textbf{41.15} \\
		\midrule
		HALO \cite{ref14}       & 24.76 & 23.79 & 38.31 & 34.10 \\
		Ours (HALO)             & \textbf{29.95} & \textbf{27.17} & \textbf{47.30} & \textbf{44.18} \\
		\bottomrule
	\end{tabular}
\end{table}

\begin{table}[t]
\centering
\caption{Performance Comparison of Drone$\rightarrow$Satellite Task on the CVOGL Dataset (\%). Bold values indicate the best results under each backbone.}
\label{tab:comparison_drone}
\setlength{\tabcolsep}{6pt}
\renewcommand{\arraystretch}{1.3}

\begin{tabularx}{\columnwidth}{c | c c c c}

\Xhline{1.2pt}
\multirow{3}{*}[-0.9em]{{Method}}
& \multicolumn{4}{c}{Drone $\rightarrow$ Satellite} \\
\cmidrule(lr){2-5}
& \multicolumn{2}{c}{Test} & \multicolumn{2}{c}{Validation} \\
\cmidrule(lr){2-3} \cmidrule(lr){4-5}
& Acc@0.25 & Acc@0.5 & Acc@0.25 & Acc@0.5 \\
\midrule

CVM-Net \cite{ref24}        & 20.14 & 3.29  & 20.04 & 3.47  \\
RK-Net \cite{ref27}         & 19.22 & 2.67  & 19.94 & 3.03  \\
L2LTR \cite{ref26}          & 38.95 & 6.27  & 38.68 & 5.96  \\
Polar-SAFA \cite{ref20}     & 37.41 & 6.58  & 36.19 & 6.39  \\
TransGeo \cite{ref25}       & 35.05 & 6.37  & 34.78 & 5.42  \\
Sample4Geo \cite{ref17}     & 32.50 & 7.15  & 24.60 & 7.15  \\
\midrule
DetGeo \cite{ref1}          & 61.97 & 57.66 & 59.81 & 55.15 \\
DetGeo + Ours               & \textbf{69.17} & \textbf{64.65} & \textbf{68.04} & \textbf{63.49} \\
\midrule
OCGNet \cite{ref12}         & 68.35 & 63.93 & 66.52 & 61.86 \\
OCGNet + Ours               & \textbf{69.99} & \textbf{64.95} & \textbf{69.56} & \textbf{64.25} \\
\midrule
HALO \cite{ref14}           & 71.53 & 64.44 & 70.21 & 63.81 \\
HALO + Ours                 & \textbf{75.13} & \textbf{70.20} & \textbf{73.35} & \textbf{65.76} \\
\Xhline{1.3pt}
\end{tabularx}
\end{table}

\subsection{Comparison Results}
 We compare them with the following methods on the CVOGL dataset for evaluating the effectiveness of our approach, including metric learning-based approaches (CVM-Net \cite{ref24}, RK-Net \cite{ref27}, Sample4Geo \cite{ref17}), the geometry-based method Polar-SAFA \cite{ref20}, Transformer-based methods (L2LTR \cite{ref26}, TransGeo \cite{ref25}), and the three cross-view object geo-localization methods mentioned above. All comparison results follow the evaluation metrics ($\mathrm{Acc}@0.25$ and $\mathrm{Acc}@0.5$), as summarized in Table~\ref{tab:comparison_drone} (Drone$\rightarrow$Satellite) and Table~\ref{tab:comparison_ground} (Ground$\rightarrow$Satellite).

\subsubsection{Quantitative Results}

\textbf{Evaluation on New Multi-Object Test Sets.}
To more rigorously assess the prompt-following capability of CVOGL models, we further evaluate them on the newly constructed multi-object test sets. Unlike the original test set where each image contains only one annotated object, the new test sets contain multiple annotated objects per image, requiring models to correctly distinguish and localize different targets based on different click prompts. Table~\ref{tab:new_test} reports the results of all evaluated models on both new test sets.

As shown in Table~\ref{tab:new_test}, all models experience a significant performance drop on the new multi-object test sets compared to the original test set, confirming the increased difficulty of the multi-object setting. Our approach consistently improves all evaluated backbones on both the G$\rightarrow$S and D$\rightarrow$S tasks. Specifically, it improves DetGeo by 3.50\% and 3.14\% in $\mathrm{Acc}@0.25$ and $\mathrm{Acc}@0.5$ on the G$\rightarrow$S test set, and brings gains of 12.01\% and 11.83\% on the D$\rightarrow$S setting. When applied to the OCGNet backbone, our approach achieves gains of 5.32\% and 4.34\% on G$\rightarrow$S, with improvements of 8.99\% and 9.07\% under the D$\rightarrow$S protocol. For the HALO backbone, it yields improvements of 5.19\% and 3.38\% on G$\rightarrow$S and obtains 8.99\% and 10.08\% gains on D$\rightarrow$S. These results demonstrate that CVOGL-MO and CVMCL effectively enhance the model's ability to follow prompt instructions, rather than merely memorizing the originally annotated object.

\noindent \textbf{Evaluation on Original Test Sets.}
As shown in Table~\ref{tab:comparison_drone}, our approach achieves consistent accuracy improvements across all baseline models on the D$\rightarrow$S task. Specifically, it improves DetGeo by 7.20\% and 6.99\% in $\mathrm{Acc}@0.25$ and $\mathrm{Acc}@0.5$. When integrated with the OCGNet backbone, our approach obtains gains of 1.64\% and 1.02\%, and for the HALO baseline it delivers improvements of 3.60\% and 5.76\% on the D$\rightarrow$S test set.

Notably, after integrating CVOGL-MO and CVMCL, the accuracy of the G$\rightarrow$S task on the original test set decreases for all three baselines while D$\rightarrow$S consistently benefits. We analyze this phenomenon in detail below.

\begin{table}[t]
	\centering
	\caption{Performance Comparison of Ground$\rightarrow$Satellite Task on the CVOGL Dataset (\%). The accuracy change is analyzed in Sec.~\ref{sec:why_g2s}.}
	\label{tab:comparison_ground}
	\setlength{\tabcolsep}{6pt}
	\renewcommand{\arraystretch}{1.3}

	\begin{tabularx}{\columnwidth}{c | c c c c}

		\Xhline{1.2pt}
		\multirow{3}{*}[-0.9em]{{Method}}
		& \multicolumn{4}{c}{Ground $\rightarrow$ Satellite} \\
		\cmidrule(lr){2-5}
		& \multicolumn{2}{c}{Test} & \multicolumn{2}{c}{Validation} \\
		\cmidrule(lr){2-3} \cmidrule(lr){4-5}
		& Acc@0.25 & Acc@0.5 & Acc@0.25 & Acc@0.5 \\
		\midrule

		DetGeo \cite{ref1}          & 45.43 & 42.24 & 46.70 & 43.99 \\
		DetGeo + Ours               & 39.15 & 36.38 & 42.04 & 39.22 \\
		\midrule
		OCGNet \cite{ref12}         & 51.49 & 47.69 & 48.54 & 44.20 \\
		OCGNet + Ours               & 46.87 & 43.78 & 44.20 & 41.06 \\
		\midrule
		HALO \cite{ref14}           & 51.59 & 46.76 & 50.81 & 46.91 \\
		HALO + Ours                 & 47.07 & 43.06 & 47.45 & 43.23 \\
		\Xhline{1.3pt}
	\end{tabularx}
\end{table}

\subsubsection{Why does G$\rightarrow$S lose?}
\label{sec:why_g2s}
We first identify which component causes the degradation through controlled experiments, and then provide evidence for the underlying cause.

\textbf{Component attribution.} As shown in the G$\rightarrow$S columns of Table~\ref{tab:ablation}, training with CVOGL-MO alone changes the original-test accuracy by $-1.64$, $-6.17$, and $-5.87$ points for DetGeo, OCGNet, and HALO, respectively, while CVMCL alone changes it by $+3.80$, $-2.98$, and $+1.44$ points. This indicates that the degradation is mainly attributable to the CVOGL-MO dataset rather than to the CVMCL loss. Specifically, the multi-object annotations of CVOGL-MO alter the training distribution so that the model is optimized to discriminate multiple objects per scene, which conflicts with the single-object setting of the original G$\rightarrow$S test set and lowers its accuracy there; CVMCL alone, in contrast, has a much smaller and even positive effect on the original test set.

\textbf{Task difficulty.} For the D$\rightarrow$S task, the cross-view gap is relatively small, and the model has sufficient capacity to learn genuine prompt-conditional localization once the shortcut is removed, yielding consistent gains. For the G$\rightarrow$S task, the severe viewpoint discrepancy makes prompt-conditional learning substantially harder, so breaking the shortcut temporarily degrades the original-test accuracy.

\textbf{Evidence for the shortcut explanation.} Our hypothesis is that the original benchmark inadvertently rewards prompt-disobeying behavior: with only 4,343 training pairs, a model that memorizes the annotated object of each scene is barely penalized on the original test set. To test this hypothesis independently of our method, we compare the baseline accuracy under the original click (C0) with that under the removed prompt (C2). As shown in Table~\ref{tab:prompt_cond_new}, on the new multi-object test sets the two are nearly identical on the G$\rightarrow$S task, e.g., DetGeo achieves 20.77\% vs.\ 21.74\%, OCGNet 22.71\% vs.\ 21.50\%, and HALO 23.79\% vs.\ 22.83\%. This near-invariance of accuracy to prompt removal directly supports the shortcut hypothesis: the baselines memorize the annotated object of each scene and do not rely on the prompt to localize the target.

\subsubsection{Qualitative Visualization and Analysis}
\begin{figure*}
	\centering
	\includegraphics[width=\linewidth]{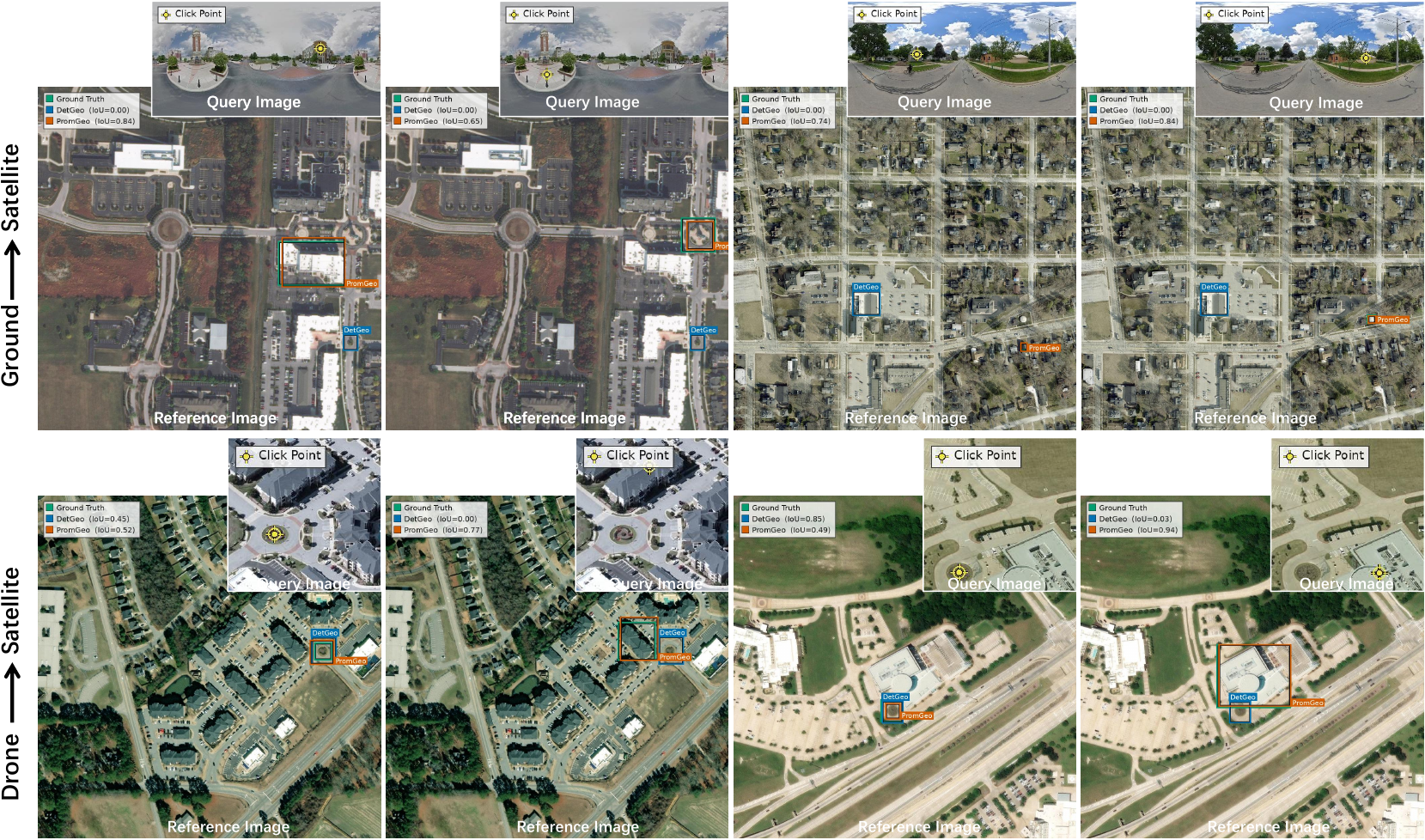}
	\caption{Visualization of some representative results of DetGeo and our approach (CVOGL-MO + CVMCL).}
	\label{fig:vis1}
\end{figure*}

\begin{figure*}
	\centering
	\includegraphics[width=\linewidth]{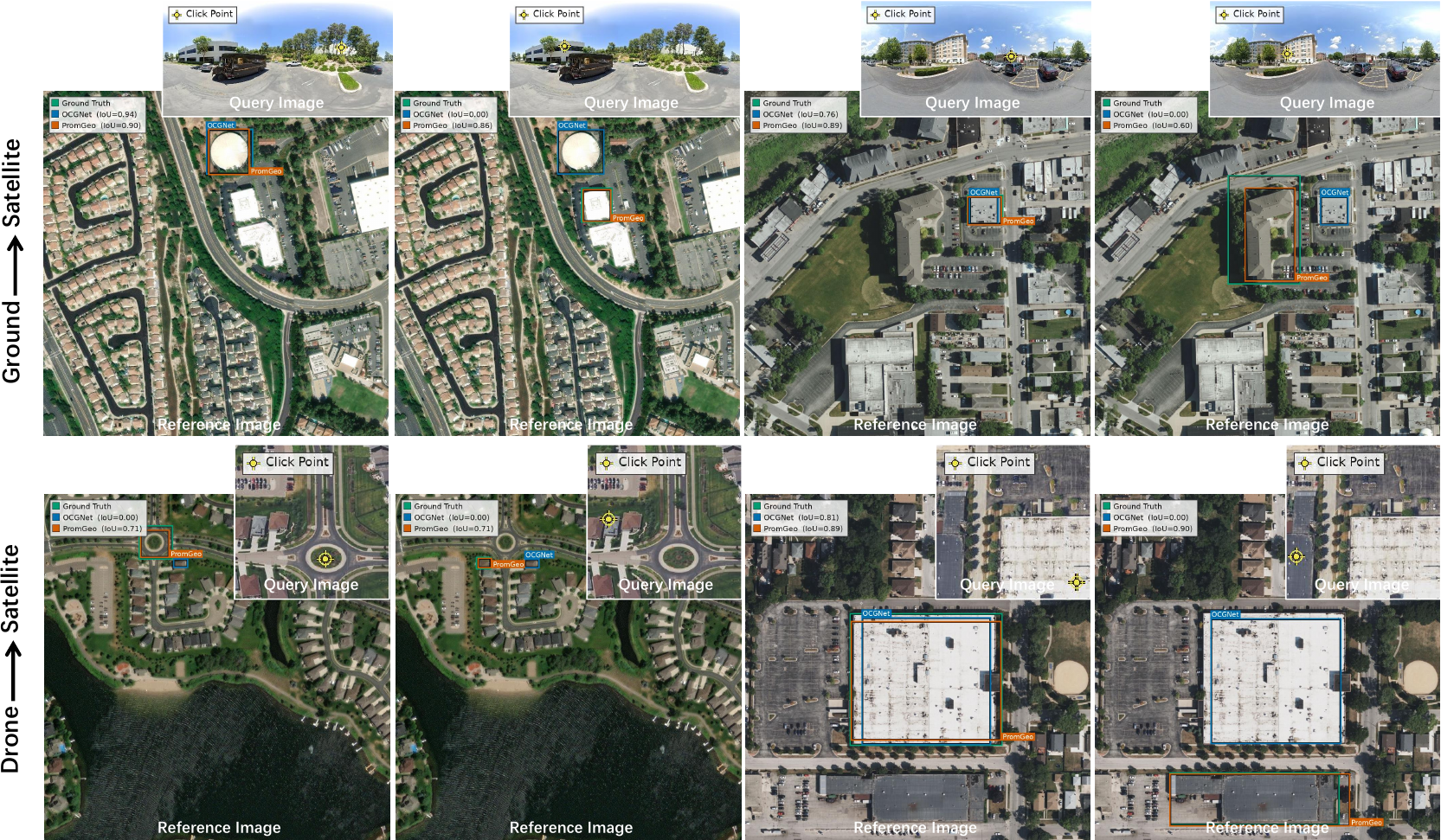}
	\caption{Visualization of some representative results of OCGNet and our approach (CVOGL-MO + CVMCL).}
	\label{fig:vis2}
\end{figure*}

\begin{figure*}
	\centering
	\includegraphics[width=\linewidth]{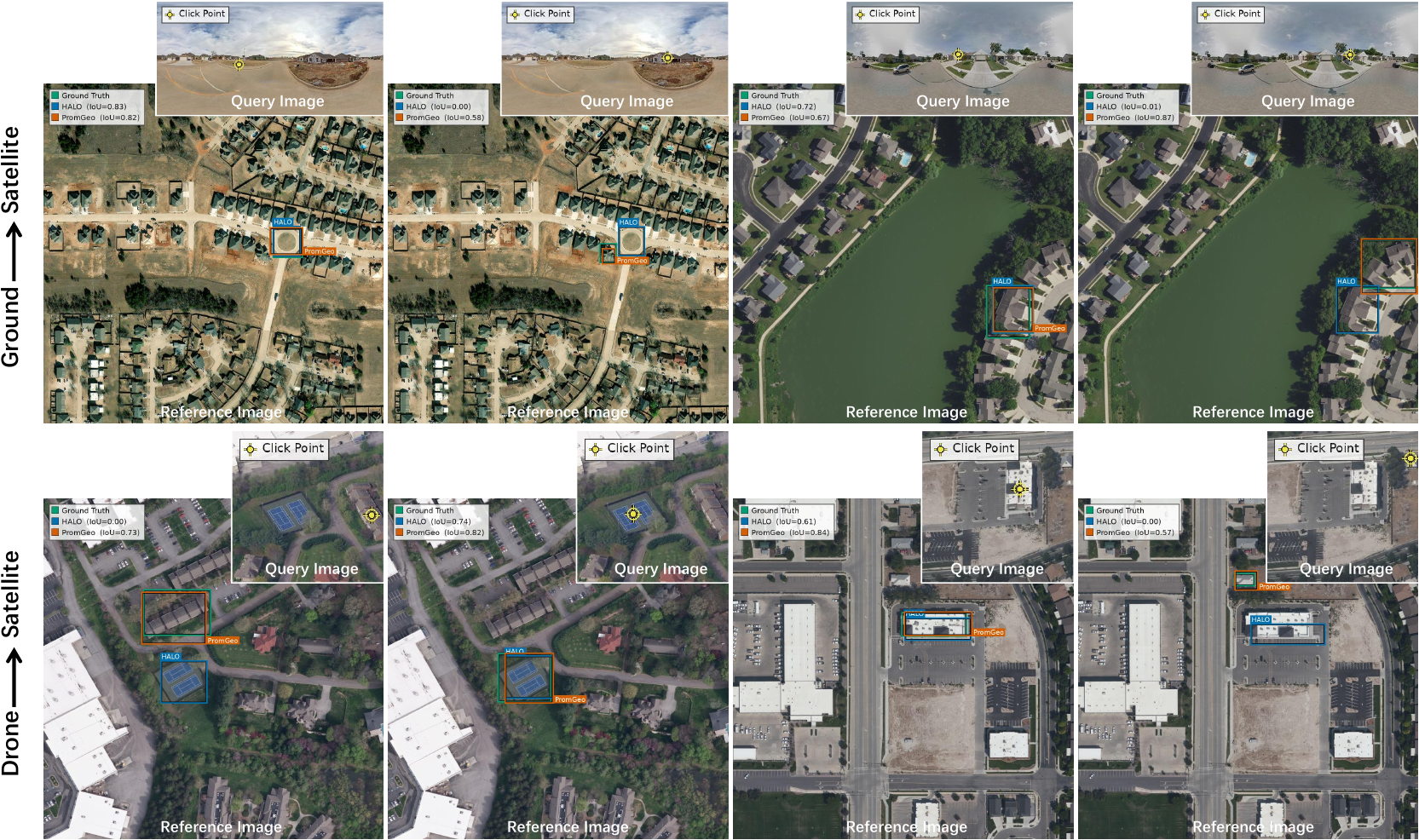}
	\caption{Visualization of some representative results of HALO and our approach (CVOGL-MO + CVMCL).}
	\label{fig:vis3}
\end{figure*}

To intuitively demonstrate the localization precision of our approach, we provide qualitative comparisons with DetGeo, OCGNet and HALO. Fig.~\ref{fig:vis1}, Fig.~\ref{fig:vis2} and Fig.~\ref{fig:vis3} provide the localization results in some representative samples of the new test set.

As shown in the three figures, when the click point is moved to a different object within the same scene, the prediction boxes of the baseline methods (DetGeo, OCGNet, and HALO) remain anchored to the initially annotated object and fail to respond to the prompt change, clearly exhibiting prompt-disobeying behavior. In contrast, models trained with CVOGL-MO and CVMCL correctly interpret the semantic intent of each click point and dynamically adjust their predictions to localize the intended object.


\subsection{Ablation Study}
We conduct thorough ablation studies on the three baselines to evaluate the contribution of each core component. The results are summarized in Table~\ref{tab:ablation}.

\begin{table*}[t]
\centering
\caption{Ablation Study Results (Acc@0.5, \%) on the Original CVOGL Test Sets of Both Tasks.}
\label{tab:ablation}
\setlength{\tabcolsep}{4pt}
\begin{tabular}{l cc | ccc | ccc}
\toprule
\multirow{2}{*}[-0.5em]{Method} & \multirow{2}{*}[-0.5em]{\makecell{CVOGL\\-MO}} & \multirow{2}{*}[-0.5em]{CVMCL}
& \multicolumn{3}{c|}{Drone $\rightarrow$ Satellite} & \multicolumn{3}{c}{Ground $\rightarrow$ Satellite} \\
\cmidrule(lr){4-6} \cmidrule(lr){7-9}
& & & DetGeo & OCGNet & HALO & DetGeo & OCGNet & HALO \\
\midrule
Baseline            & $\times$      & $\times$      & 57.66 & 63.93 & 64.44 & 42.24 & 47.69 & 46.76 \\
w/ CVOGL-MO         & $\checkmark$  & $\times$      & 62.59 & 64.34 & 69.78 & 40.60 & 41.52 & 40.89 \\
w/ CVMCL            & $\times$      & $\checkmark$  & 57.35 & 62.38 & 66.29 & 46.04 & 44.71 & 48.20 \\
\midrule
CVOGL-MO + CVMCL    & $\checkmark$  & $\checkmark$  & \textbf{64.65} & \textbf{64.95} & \textbf{70.20} & 41.34 & 43.78 & 43.06 \\
\bottomrule
\end{tabular}
\end{table*}

\begin{table}[t]
	\centering
	\caption{Ablation Study Results of DetGeo on the New CVOGL-MO Test Set (\%).}
	\label{tab:ablation1}
	\setlength{\tabcolsep}{1pt}
	\renewcommand{\arraystretch}{1.2}
	\begin{tabular}{c | c | c | cc | cc}
		\toprule
		\multicolumn{3}{c|}{Components}
		& \multicolumn{2}{c|}{Ground $\rightarrow$ Satellite}
		& \multicolumn{2}{c}{Drone $\rightarrow$ Satellite} \\
		\midrule
		DetGeo & CVOGL-MO & CVMCL & Acc@0.25 & Acc@0.5 & Acc@0.25 & Acc@0.5 \\
		\midrule
		 $\checkmark$ & $ $ & $ $ & 21.74 & 20.77 & 27.59 & 25.57 \\
		 $\checkmark$ & $\checkmark$ & $ $ & 24.65 & 23.69 & 39.51 & 37.21 \\
		 $\checkmark$ & $ $ & $\checkmark$ & 23.91 & 22.71 & 30.61 & 28.60 \\
		 \midrule
		 $\checkmark$ & $\checkmark$ & $\checkmark$ & \textbf{25.24} & \textbf{23.91} & \textbf{39.60} & \textbf{37.40} \\
		\bottomrule
	\end{tabular}
\end{table}

\begin{table}[t]
	\centering
	\caption{Ablation Study Results of OCGNet on the New CVOGL-MO Test Set (\%).}
	\label{tab:ablation2}
	\setlength{\tabcolsep}{1pt}
	\renewcommand{\arraystretch}{1.2}
	\begin{tabular}{c | c | c | cc | cc}
		\toprule
		\multicolumn{3}{c|}{Components}
		& \multicolumn{2}{c|}{Ground $\rightarrow$ Satellite}
		& \multicolumn{2}{c}{Drone $\rightarrow$ Satellite} \\
		\midrule
		OCGNet & CVOGL-MO & CVMCL & Acc@0.25 & Acc@0.5 & Acc@0.25 & Acc@0.5 \\
		\midrule
		$\checkmark$ & $ $ & $ $ & 24.03 & 22.71 & 35.01 & 32.08 \\
		$\checkmark$ & $\checkmark$ & $ $ & 27.29 & 25.36 & 42.80 & 40.51 \\
		$\checkmark$ & $ $ & $\checkmark$ & 23.55 & 22.22 & 34.83 & 31.71 \\
		\midrule
		$\checkmark$ & $\checkmark$ & $\checkmark$ & \textbf{29.35} & \textbf{27.05} & \textbf{44.00} & \textbf{41.15} \\
		\bottomrule
	\end{tabular}
\end{table}

\begin{table}[t]
	\centering
	\caption{Ablation Study Results of HALO on the New CVOGL-MO Test Set (\%).}
	\label{tab:ablation3}
	\setlength{\tabcolsep}{1pt}
	\renewcommand{\arraystretch}{1.2}
	\begin{tabular}{c | c | c | cc | cc}
		\toprule
		\multicolumn{3}{c|}{Components}
		& \multicolumn{2}{c|}{Ground $\rightarrow$ Satellite}
		& \multicolumn{2}{c}{Drone $\rightarrow$ Satellite} \\
		\midrule
		HALO & CVOGL-MO & CVMCL & Acc@0.25 & Acc@0.5 & Acc@0.25 & Acc@0.5 \\
		\midrule
		$\checkmark$ & $ $ & $ $ & 24.76 & 23.79 & 38.31 & 34.10 \\
		$\checkmark$ & $\checkmark$ & $ $ & 29.71 & 26.96 & 45.46 & 42.35 \\
		$\checkmark$ & $ $ & $\checkmark$ & 26.09 & 24.28 & 39.05 & 35.56 \\
		\midrule
		$\checkmark$ & $\checkmark$ & $\checkmark$ & \textbf{29.95} & \textbf{27.17} & \textbf{47.30} & \textbf{44.18} \\
		\bottomrule
	\end{tabular}
\end{table}

\noindent \textbf{Effectiveness of CVOGL-MO Dataset.}
We first isolate the effect of the CVOGL-MO dataset by training baselines solely on the expanded annotations without CVMCL. On the original D$\rightarrow$S test set (Table~\ref{tab:ablation}), CVOGL-MO alone brings gains of 4.93\% for DetGeo, 0.41\% for OCGNet, and 5.34\% for HALO (Acc@0.5). The trend is more pronounced on the new multi-object test sets (Tables~\ref{tab:ablation1}--\ref{tab:ablation3}): for the D$\rightarrow$S task, CVOGL-MO alone delivers improvements of 11.92\%, 7.79\%, and 7.15\% in Acc@0.25 for DetGeo, OCGNet, and HALO, respectively, and it consistently improves all three backbones on the G$\rightarrow$S task as well.
These results confirm that the diversified annotations compel the model to discriminate between multiple objects within the same scene, directly mitigating the prompt-disobeying shortcut.

\noindent \textbf{Effectiveness of CVMCL.}
We then evaluate CVMCL in isolation, i.e., training baselines with CVMCL on the original CVOGL dataset (without multi-object expansion). On the original D$\rightarrow$S test set (Table~\ref{tab:ablation}), CVMCL alone slightly changes the accuracy of the baselines (e.g., $-0.31\%$ for DetGeo and $-1.55\%$ for OCGNet, with a modest $+1.85\%$ for HALO), and on the new multi-object test sets its individual gains remain limited compared with CVOGL-MO alone. This is expected: CVMCL is specifically designed for multi-object scenes, where it separates the features of different click targets within the same scene. When applied to the original single-object dataset, each scene contains only one annotated object, so the co-scene contrastive groups are sparse and the loss has little signal to exploit. Consequently, adding CVMCL alone yields only a marginal improvement on the single-object baselines, and its full benefit is realized only when paired with the multi-object annotations of CVOGL-MO.

\noindent \textbf{Complementarity of CVOGL-MO and CVMCL.}
The full combination of CVOGL-MO and CVMCL achieves the best results across settings, and the two components exhibit clear complementarity. For example, on the D$\rightarrow$S new test set, DetGeo achieves 39.51\% with CVOGL-MO alone and 30.61\% with CVMCL alone, while the combination reaches 39.60\%; the same holds for OCGNet (42.80\% and 34.83\% individually vs.\ 44.00\% combined) and HALO (45.46\% and 39.05\% individually vs.\ 47.30\% combined). This pattern holds across all backbones: the CVOGL-MO dataset provides the necessary multi-object training signal at the data level, while CVMCL provides the complementary optimization-level constraint that explicitly separates the features of different prompted objects within the same scene. The two components address the problem from complementary directions, and their combination yields the most robust prompt-consistent geo-localization.

\begin{figure}
	\centering
	\includegraphics[width=\linewidth]{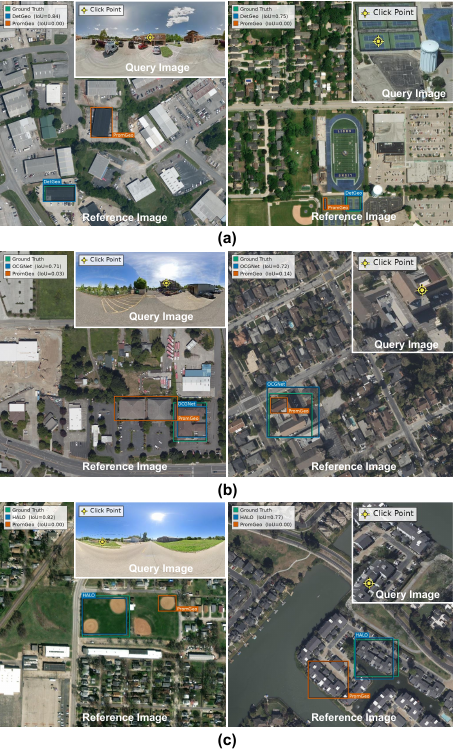}
	\caption{Representative failure cases of our approach integrated with DetGeo, OCGNet, and HALO.}
	\label{fig:fail}
\end{figure}

\section{Discussion}
\label{sec:discussion}
Although our approach consistently improves prompt-consistency across all backbones, the absolute accuracy on the multi-object test sets remains limited. Fig.~\ref{fig:fail} presents representative failure cases of our approach integrated with DetGeo, OCGNet, and HALO. Notably, in all three cases the baseline methods (DetGeo, OCGNet, and HALO) make correct predictions, while our approach fails.

A common pattern across all failure cases is that the prompted target has one or more visually near-identical counterparts in the same satellite scene. In Fig.~\ref{fig:fail}(a), for the G$\rightarrow$S task, our approach confuses the prompted house with a house sharing the same black roof; for the D$\rightarrow$S task, the ground-truth annotation box contains two rectangular blue sports courts, yet several other visually identical blue courts exist nearby, and our approach incorrectly localizes one of the similar courts instead of the prompted one. In Fig.~\ref{fig:fail}(b), our approach selects a neighboring house that closely resembles the actual target in appearance in the G$\rightarrow$S case, and in the D$\rightarrow$S case it captures only one wing of an irregularly shaped (gate-shaped) building, failing to encompass the complete structure. In Fig.~\ref{fig:fail}(c), our approach is misled by a visually similar green lawn adjacent to the true target in G$\rightarrow$S, and by another house of similar appearance in D$\rightarrow$S.

Beyond visual similarity, irregular object shapes further compound the difficulty. Several targets in the dataset are entire building complexes with non-rectangular footprints (e.g., gate-shaped or L-shaped structures). The model often predicts only a sub-region of such objects. We have unified the boundary conventions for such cases in CVOGL-MO (Sec.~III-B); nevertheless, inferring the full extent of irregularly shaped targets from a single click remains challenging.

\section{Conclusion}
In this work, we identify and quantitatively characterize the prompt-disobeying overfitting problem in cross-view object geo-localization. Using a dedicated prompt-sensitivity evaluation protocol, we find that existing models show little sensitivity to click prompt variations. Owing to the limited benchmark size, these models overfit and memorize fixed responses, failing to adhere to prompt instructions. To overcome this, we introduce two plug-and-play components: the CVOGL-MO dataset, which enlarges the training data with multi-object annotations, and the CVMCL loss, which enforces intra-scene feature discrimination between different click prompts. Experiments on DetGeo, OCGNet, and HALO demonstrate that these components improve prompt-following accuracy on the new multi-object test sets and markedly increase prompt sensitivity, without any architectural modification.

\end{document}